\documentclass[conference]{IEEEtran}
\IEEEoverridecommandlockouts
\usepackage{cite}
\usepackage{amsmath,amssymb,amsfonts}
\usepackage{algorithmic}
\usepackage{graphicx}
\usepackage{textcomp}
\usepackage{xcolor}
\def\BibTeX{{\rm B\kern-.05em{\sc i\kern-.025em b}\kern-.08em
    T\kern-.1667em\lower.7ex\hbox{E}\kern-.125emX}}
\usepackage{adjustbox}
\usepackage{microtype}
\usepackage{booktabs}
\usepackage{graphicx}
\usepackage{subcaption}
\usepackage{enumitem}
\usepackage{multirow}
\usepackage{makecell}

\usepackage{subeqnarray}
\usepackage{algorithm}
\usepackage{algorithmic}
\usepackage{amsthm}
\DeclareMathOperator*{\argmax}{arg\,max}

\usepackage[hidelinks]{hyperref}
\theoremstyle{plain}

\theoremstyle{definition}

\theoremstyle{remark}

\begin{document}

\title{
    Low-Resource Named Entity Recognition: Can One-vs-All AUC Maximization Help?
}
\author{\IEEEauthorblockN{
Ngoc Dang Nguyen
}
\IEEEauthorblockA{
\textit{Department of Data Science and AI} \\
\textit{Monash University}\\
Melbourne, Australia \\
dan.nguyen2@monash.edu
}
\and
\IEEEauthorblockN{
Wei Tan
}
\IEEEauthorblockA{
\textit{Department of Data Science and AI} \\
\textit{Monash University}\\
Melbourne, Australia \\
wei.tan2@monash.edu
}
\and
\IEEEauthorblockN{Lan Du*\thanks{* Corresponding Author}}
\IEEEauthorblockA{
\textit{Department of Data Science and AI} \\
\textit{Monash University}\\
Melbourne, Australia \\
lan.du@monash.edu
}\and
\IEEEauthorblockN{Wray Buntine}
\IEEEauthorblockA{
\textit{College of Engineering and Computer Science} \\
\textit{VinUniversity}\\
Hanoi, Vietnam \\
wray.b@vinuni.edu.vn
}
\and
\IEEEauthorblockN{Richard Beare}
\IEEEauthorblockA{
\textit{Faculty of Medicine}\\
\textit{Peninsula Clinical School} \\
Melbourne, Australia \\
richard.beare@monash.edu
}
\and
\IEEEauthorblockN{Changyou Chen}
\IEEEauthorblockA{
\textit{Department of CS and Engineering} \\
\textit{University at Buffalo}\\
New York, USA \\
changyou@buffalo.edu
}
}

\maketitle

\IEEEpeerreviewmaketitle 

\begin{abstract}
Named entity recognition (NER), a task that identifies and categorizes named entities such as persons or organizations from text, is traditionally framed as a multi-class classification problem. However, this approach often overlooks the issues of imbalanced label distributions, particularly in low-resource settings, which is common in certain NER contexts, like biomedical NER (bioNER). To address these issues, we propose an innovative reformulation of the multi-class problem as a one-vs-all (OVA) learning problem and introduce a loss function based on the area under the receiver operating characteristic curve (AUC). To enhance the efficiency of our OVA-based approach, we propose two training strategies: one groups labels with similar linguistic characteristics, and another employs meta-learning. 
The superiority of our approach is confirmed by its performance, which surpasses
traditional NER learning
in varying NER settings.
\end{abstract}


\section{Introduction}

Named Entity Recognition (NER), a fundamental task in Natural Language Processing (NLP), aims to detect the semantic category of named entities (NE), such as location, organization, or person. As an essential prerequisite for many language applications, NER often plays a pivotal role in a variety of NLP tasks, including information extraction, information retrieval, task-oriented dialogues, and knowledge base construction \cite{devlin-etal-2019-bert,li2020survey}. NER's relevance extends to numerous real-world applications; for instance, it aids in extracting information from unstructured text in the medical domain to improve patient care, it enhances search engine algorithms to better understand user queries, and it supports intelligence services by identifying crucial entities in large text corpora \cite{li2020survey}.

Recently, NER has seen significant performance improvements due to the advances of state-of-the-art (SOTA) pre-trained language models (PLMs) \cite{devlin-etal-2019-bert}. However, these PLMs rely heavily on sizable training datasets, and in specialized low-resource settings ({\it e.g.}, biomedical domains), the lack of data often leads to sub-optimal performance \cite{yaseen2021data}. This poses a significant challenge, as many languages and specialized domains suffer from a lack of annotated data \cite{li2020survey,nguyen2022auc}. Improving NER performance in these low-resource settings has the potential to democratize access to advanced language processing capabilities, thus breaking down barriers to information access and enabling new NLP applications across a multitude of disciplines and industries.

\begin{figure}[t]
\centering
    \begin{subfigure}[t]{0.225\textwidth}
        \centering
        \includegraphics[width=\textwidth]{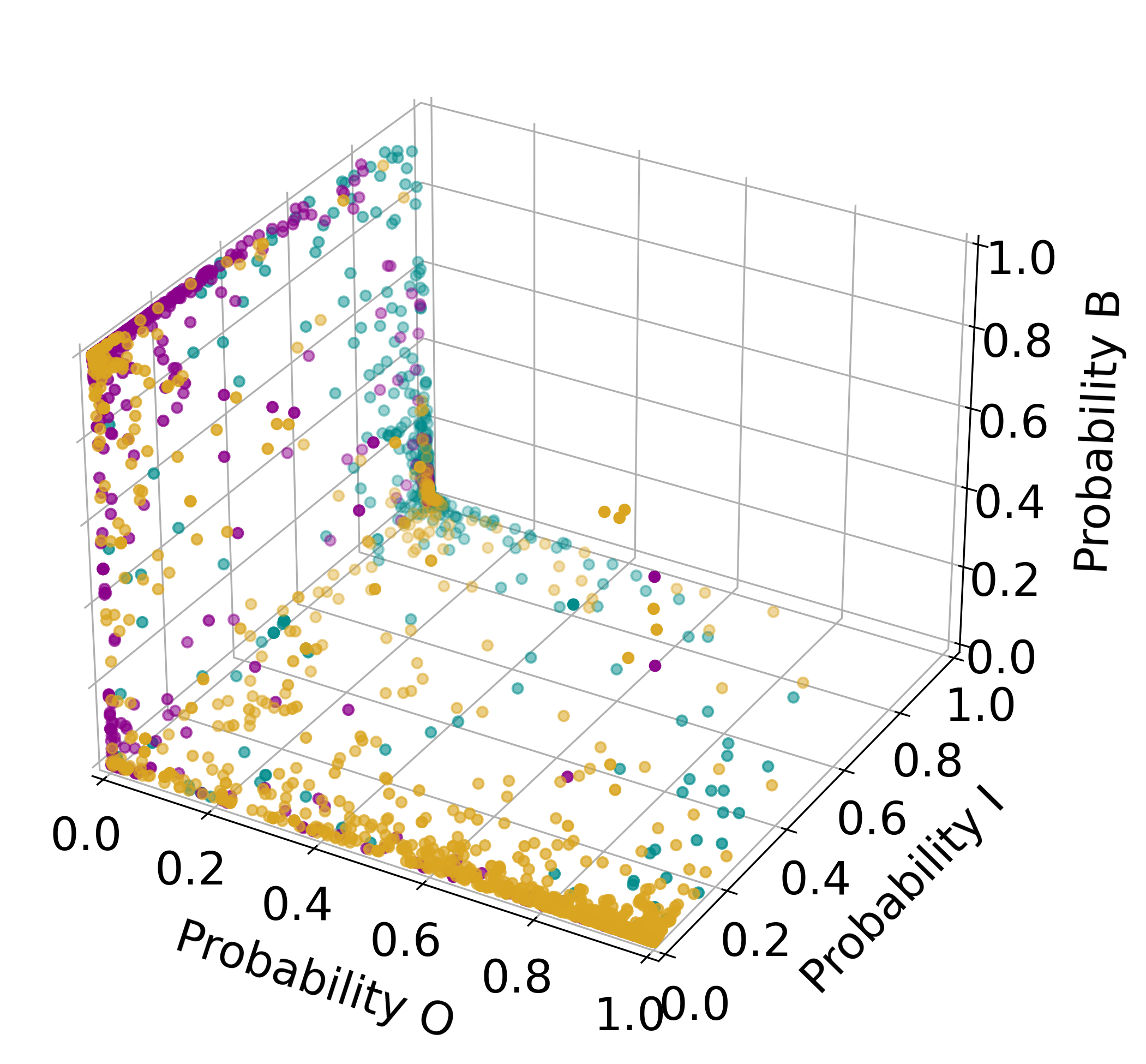}
        \caption{AUC}
        \label{fig:AUC_3d}
    \end{subfigure}
    \hfill
    \begin{subfigure}[t]{0.225\textwidth}
        \centering
        \includegraphics[width=\textwidth]{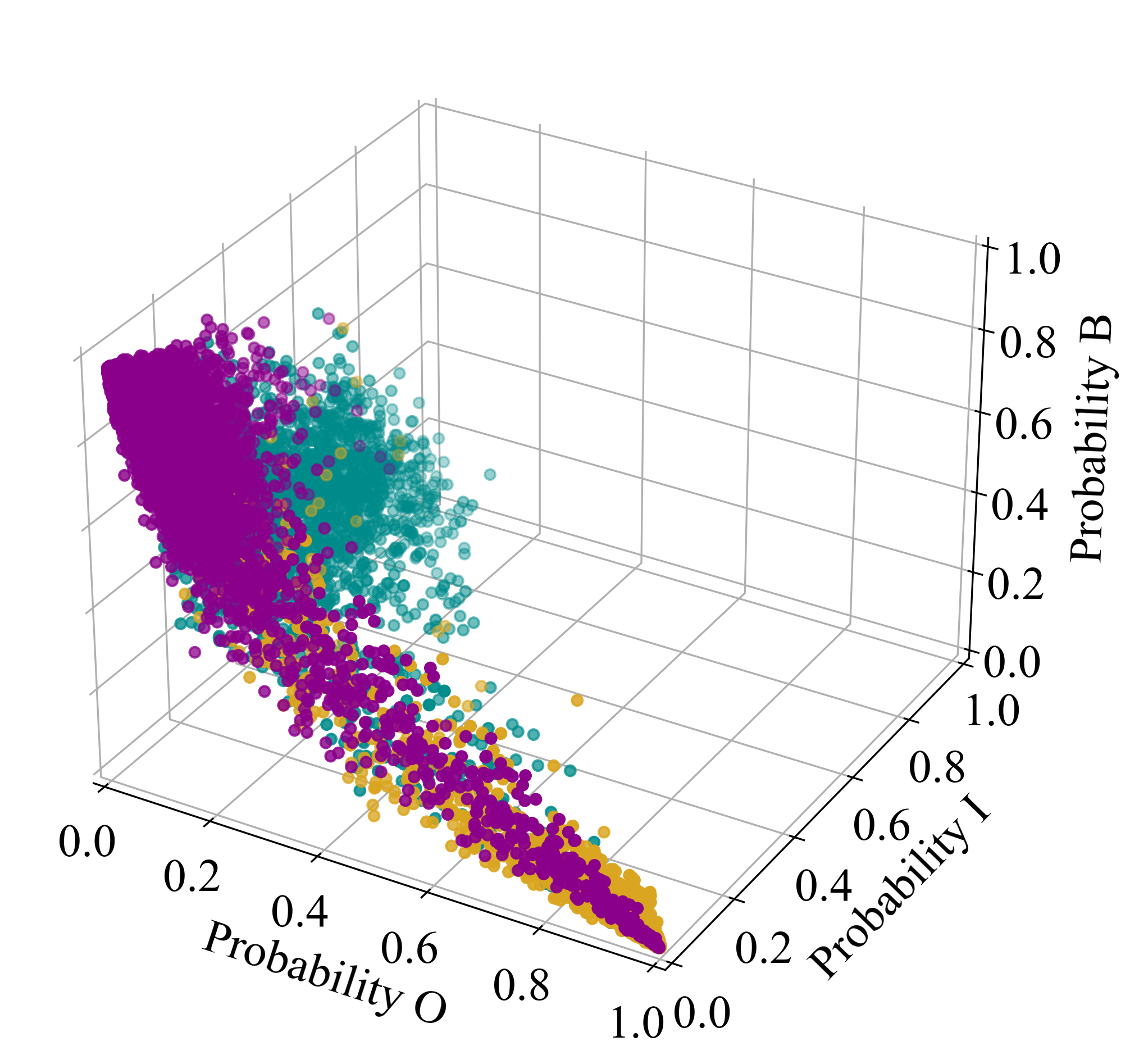}
        \caption{BCE}
        \label{fig:BCE_3d}
    \end{subfigure}
    \begin{subfigure}[t]{0.5\textwidth}
        \centering
        \includegraphics[width=1\textwidth]{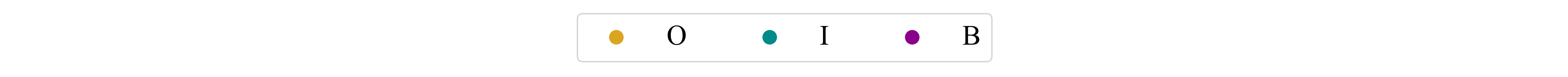}
        \label{fig:OVA_AUC_BCE_legend}
    \end{subfigure}
    \caption{
        Illustration of the OVA predicted probability performance of classifiers trained with AUC and BCE losses for each token's BIO-tag on the CoNLL 2003 test data.
        The binary classifiers trained with the BCE loss exhibit a lack of confidence in token classification, as indicated by the clustering of all predictions rather than their stratification towards their respective label corners. Conversely, when trained with the AUC loss, these binary classifiers effectively classify tokens, showing the potential of AUC for enhanced NER performance.
    }
    \label{fig:OVA_BCE_3d_comparison}
\end{figure}

NER models are often heavily dependent on human-annotated data, which can be costly and time-consuming to obtain. To alleviate this dependence on labeled data, existing low-resource machine learning approaches such as domain adaptation \cite{nguyen-etal-2022-hardness}, and data augmentation \cite{zhou2022melm} have been applied to NER. However, these methods do not account for the imbalanced label distributions commonly found in NER datasets, where the majority of labels in the NER corpora are of the non-entity type ``O'' (see \autoref{tab:DatasetSetting_BIO}). These non-entity labels provide limited learning signals for NER models, which presents challenges to the NER classifier in correctly identifying NEs in sentences. This imbalance is particularly prominent in specialized biomedical corpora, such as NCBI \cite{10.5555/2598938.2599127}, and s800 \cite{10.1371/journal.pone.0065390}, where over 90\% of labels are tagged as ``O'' (see \autoref{tab:DatasetSetting_BIO}).

Given the inherent imbalanced label distributions in NER corpora, we argue that even though standard learning objectives ({\it i.e.}, cross entropy loss or conditional random fields), given adequate annotated training data, are capable of producing well-performing token classifiers or sequence labellers, their performance can substantially degrade in low-resource settings. Consequently, we propose to directly address the imbalance issue by training NER models with a surrogate loss that maximizes the area under the ROC curve (AUC). AUC maximization has been demonstrated to greatly improve model prediction for tasks with imbalanced label distribution \cite{gao2013one, ying2016stochastic,yuan2020large}. In the NER task, the recent work of \cite{nguyen2022auc} has attempted to utilize AUC maximization by reformulating the NER task as a two-task learning problem. While this approach shows significant improvements over traditional methods, it fails to directly address the problem of predicting entity types, such as location, organization, or person, limiting its usage in real-world applications where identifying the entity type is crucial.

Thus, multi-class AUC objectives, which cannot be developed using the aforementioned two-task approach \cite{nguyen2022auc}, are essential for NER. They can be developed with two strategies. The first is the one-vs-other (OVO) reformulation \cite{yang2021learning}, which is not computationally feasible due to its quadratic relationship with the number of classes. Consequently, we reformulate the standard sequence tagging problem as a one-vs-all (OVA) learning problem. In this context, each unique label ({\it e.g.}, B-PER, I-ORG, O, {\it etc.}) will have its own binary classifier, which can be individually learned with an AUC objective function.

However, OVA has two notable weaknesses: {\bf (i)} compared to the traditional multi-class setup, OVA has a longer training time since each binary classifier must be independently trained, and {\bf (ii)} OVA is not data efficient for learning from imbalanced data which results in less accurate classifiers compared to its multi-class counterpart \cite{liu2017action}. As shown in \autoref{fig:OVA_BCE_3d_comparison}, the predictive confidence of OVA, learned with the binary cross entropy loss (BCE), is not well-stratified compared to the predictive confidence learned with AUC. To alleviate the first weakness, we propose two new training strategies: one that groups labels sharing similar linguistic characteristics and trains their classifiers together, and another that adapts the idea of meta-learning \cite{finn2017model} by selecting a random batch of binary classifiers for the model to learn. For the second weakness, the binary classifier is tuned with an AUC surrogate loss function, which has shown its resilience to imbalanced data \cite{yuan2021compositional,yuan2020large}.

To demonstrate the effectiveness of our proposed method, we conduct thorough empirical investigations to evaluate the efficacy of our OVA-AUC-NER techniques under various scenarios. The results of the experiments, which are provided in \autoref{sec:ExpRes}, show that our techniques exhibit significant performance improvement compared to
traditional baselines in terms of entity-F1 score across corpora ({\it e.g.}, OntoNotes5, CoNLL 2003, NCBI, and s800), domains ({\it e.g.}, biomedical,general), and label distributions ({\it e.g.}, NEs percentages of $1,2,$ or $5$\%) in low-resource conditions ({\it e.g.}, training pool size of \{$20,50,100,200,300,400,500$\}). 
Our contributions include:
\begin{table*}[!t]
\centering
\small
\caption{
    Summary of dataset. Please note that we use BIO here instead of the entity type label for ease of reference. 
}
\begin{adjustbox}{max width=\textwidth}
\begin{tabular}{llcllclll}
    \toprule
    \multicolumn{1}{l}{\bf Dataset}
        & \multicolumn{1}{l}{\bf Domain}
        & \multicolumn{1}{c}{\bf \% Training}
        & \multicolumn{1}{c}{\bf \# Sentences}
        & \multicolumn{1}{c}{\bf \# Tokens}
        & \multicolumn{1}{c}{\bf \# Labels}
        & \multicolumn{3}{c}{\bf \% label (B/I/O)} \\
   \multicolumn{1}{l}{}
        & \multicolumn{1}{l}{}
        & \multicolumn{1}{c}{\bf Gold Samples}
        & \multicolumn{1}{c}{Train/Dev/Test}
        & \multicolumn{1}{c}{Train/Dev/Test}
        & {}  
        & \multicolumn{1}{c}{Train} 
        & \multicolumn{1}{c}{Dev}
        & \multicolumn{1}{c}{Test} \\
    \cmidrule(r){1-9}
    \multicolumn{1}{l}{OntoNotes5 \cite{weischedel2011ontonotes}}
        & \multicolumn{1}{l}{General}
        & {45.59}
        & {20,000}/{3,000}/{3,000}
        & {364,344}/{54,372}/{55,754}
        & 37
      & {6.2}/{4.8}/{89.0} & {6.2}/{4.8}/{89.0} & {6.1}/{4.9}/{89.0} \\ 
    \cmidrule (r){1-9}
    \multicolumn{1}{l}{CoNLL 2003 \cite{tjong-kim-sang-de-meulder-2003-introduction}}
        & \multicolumn{1}{l}{General}
        & {79.25}
        & {14,040}/{3,249}/{3,452}
        & {203,589}/{51,319}/{46,376}
        & 9
      & {11.5}/{5.2}/{83.3} & {11.6}/{5.1}/{83.3} & {12.2}/{5.3}/{82.5} \\ 
    \cmidrule (r){1-9}
    \multicolumn{1}{l}{NCBI \cite{10.5555/2598938.2599127}}
        & \multicolumn{1}{l}{Disease}
        & {53.89}
        & {5,424}/{923}/{940} 
        & {135,597}/{23,969}/{24,481}
        & 3
        & {3.8}/{4.5}/{91.7} & {3.3}/{4.5}/{92.2} & {3.9}/{4.5}/{91.6} \\
       \cmidrule (r){1-9}
    \multicolumn{1}{l}{s800 \cite{10.1371/journal.pone.0065390}}
        & \multicolumn{1}{l}{Species}
        & {29.51}
        & {5,733}/{830}/{1,630}
        & {147,205}/{22,166}/{42,287}
        & 3
        & {1.7}/{2.3}/{96.0} & {1.7}/{2.2}/{96.1} & {1.8}/{2.5}/{95.7} \\
    \bottomrule
\end{tabular}%
\end{adjustbox}
\label{tab:DatasetSetting_BIO}
\end{table*}

\begin{itemize}[topsep=0pt, partopsep=0pt, noitemsep, leftmargin=*]
    \item \textbf{Reformulation of NER as an OVA task}: We transform NER from a standard multi-class learning problem to an OVA learning problem, with each unique label having its own classifier. This simple OVA reformulation makes AUC maximization feasible for predicting the entity type in NER.
    \item \textbf{Efficient training strategies}: We propose two efficient training strategies to overcome the computational cost of training in the OVA setup; one groups labels sharing similar linguistic characteristics and trains their classifiers together while the other uses a meta-learning approach \cite{finn2017model} to randomly select a batch of classifiers for the model to learn.
\end{itemize}


\section{Related Work}
\label{sec:Background}


The OVA approach is mostly used to expand binary models for multi-class classification, such as logistic regression, support vector machines, etc \cite{GALAR20111761, 8302159}, with the OVO approach seeing little use due to its higher training time. OVA's objective is to divide a $K$-class problem into $K$ binary problems. For instance, $K$ binary classifiers must be built, where $K$ is the number of classes, and the $i$-th classifier is trained with positive data from class $i$ and negative samples from the other $K-1$ classes. When the classifier evaluates an unclassified sample, the highest confidence value of the sample is considered to have labelled corresponding to the specified class \cite{GALAR20111761}. Recently, researchers have discovered how OVA methods can accomplish various tasks with deep neural networks. They found that OVA can identify more relevant hidden representations for unidentified instances than the popular Softmax function \cite{DBLP:journals/corr/abs-2004-08067}. Additionally, OVA improves calibration on image classification, outlier detection, and dataset shift tasks, reaching Softmax's predictive performance without increasing the training or the test time complexity \cite{DBLP:journals/corr/abs-2007-05134,DBLP:journals/corr/abs-2104-03344}. Although the algorithm is simple and straightforward, it shows impressive results, demonstrating that its performance is usually at least as accurate as other multi-class algorithms when the classifiers are properly tuned \cite{rifkin2004defense}. 

AUC (Area Under the ROC Curve) has been traditionally treated as an important criterion for measuring model's classification performance \cite{freund2003efficient}. Due to its non-convex, and discontinuous nature, most works consider direct optimization of AUC score an NP-hard problem \cite{yuan2020large}. \cite{freund2003efficient} tried to alleviate this computation difficulty through a pairwise surrogate loss, while \cite{zhao2011online} implemented a hinge loss. However, both of these surrogate losses lack scalability to large datasets and models. This led to the development of the least-square surrogate loss \cite{gao2013one}. Recent research on AUC maximization further optimizes the least-square surrogate loss via deep margin surrogate loss \cite{yuan2020large} and compositional training \cite{yuan2021compositional}. Overall, AUC maximization works well when there exists an imbalanced label distribution, or the AUC score is the default metric for evaluating and comparing different methods \cite{yuan2020large}. Prior work for AUC maximization in NER includes \cite{nguyen2022auc} which fails to predict entity types, such as ORG, LOC, or PER; thus, our work is the first to explore AUC maximization for the NER task while appropriately identifying the types of the entities.


\section{AUC Maximization for NER}
\label{sec:AUCNER}

\textbf{Notation.}
Given a set of training data $\mathcal{S}=\left\{\left(\mathbf{x}_{1}, \mathbf{y}_{1}\right), \ldots,\left(\mathbf{x}_{n}, \mathbf{y}_{n}\right)\right\}$, where $\mathbf{x}_{i} = {x}_i^1, \ldots, {x}_i^l$ represents the $i$-th training example  ({\it i.e.}, a sentence of length $l$), and $\mathbf{y}_{i} \in\{\text{B}, \text{I}, \text{O}\}^{l}$ denotes its corresponding sequence of labels, we would like to learn the objective mapping function $h:\mathcal{X}\rightarrow\mathcal{Y}$. This objective mapping function, traditionally learned with either CRFs or the CE loss, is parameterized with $\mathbf{w} \in \mathbb{R}^{d}$, {\it i.e.}, $h_{\mathbf{w}}(\mathbf{x})=h(\mathbf{w}, \mathbf{x})$. Lastly, different corpora can have different label set, {\it e.g.}, ${y}_{i} \in \{\text{O}, \text{B-ORG}, \text{B-PER}, \text{B-MISC}, \text{B-LOC}, \text{I-ORG}, \text{I-PER}, \text{I-LOC},\allowbreak \text{I-MISC}\}$ for CoNLL 2003~\cite{tjong-kim-sang-de-meulder-2003-introduction}; thus, the use of $\mathbf{y}_{i} \in\{\text{B}, \text{I}, \text{O}\}^{l}$ presented in many parts of this study is for ease of reference.

To make direct AUC maximization applicable to NER, we first reformulate the standard NER multi-class setup as a one-vs-all learning problem. Consequently, each unique label is given its own binary classifier that can be learned using the AUC objective function. For instance, given the ``O''-tag, the label of this tag is $\mathbf{y}_{\textbf{O}_i}\in\{-1,1\}^{l}$, while the parameter of this task is $\mathbf{w}_{\textbf{O}} = \{\theta, \omega_{\textbf{O}}\}$. $\theta$ denotes the shared embedding and pretrained language model parameters, while $\omega_{\textbf{O}}$ denotes the parameters for the binary classifier for the ``O''-tag. After the reformulation, we can maximize the AUC score of each binary classifier via the robust deep AUC margin loss (DAM) \cite{yuan2020large}.

{\footnotesize
\begin{subeqnarray}
    \lefteqn{\text{AUC}_{\mathrm{M}}(\mathbf{w}_{\textbf{O}})}\nonumber\allowdisplaybreaks
    \\&=&\allowdisplaybreaks
    \mathbb{E}\left[\left(m_{\textbf{O}}-h_{\mathbf{w}_{\textbf{O}}}({x})+h_{\mathbf{w}_{\textbf{O}}}\left({x}^{\prime}\right)\right)^{2} \mid y_{\textbf{O}}=1, y_{\textbf{O}}^{\prime}=-1\right]
    \\&=&\allowdisplaybreaks
    \min _{a_{\textbf{O}}, b_{\textbf{O}}}{A_{1}(\mathbf{w}_{\textbf{O}})} 
    +{A_{2}(\mathbf{w}_{\textbf{O}})}
    +
    (m_{\textbf{O}}-a_{\textbf{O}}+b_{\textbf{O}})^{2}
    \\&=&\allowdisplaybreaks
    \min _{a_{\textbf{O}}, b_{\textbf{O}}}A_{1}(\mathbf{w}_{\textbf{O}})+A_{2}(\mathbf{w}_{\textbf{O}})  \nonumber
    \\&+&\allowdisplaybreaks
    \max _{\alpha_{\textbf{O}}\geq0}\left\{2 {\alpha_{\textbf{O}}}(m_{\textbf{O}}-a_{\textbf{O}}
    +b_{\textbf{O}}
    )
    -{\alpha_{\textbf{O}}}^{2}\right\},
\label{eq:AUCLoss}
\end{subeqnarray}
}where $A_{1}(\mathbf{w}_{\textbf{O}})=\mathbb{E}[h^2_{\mathbf{w}_{\textbf{O}}}({x})\mid y_{\textbf{O}}=1] - a_{\textbf{O}}^2$, $A_{2}(\mathbf{w}_{\textbf{O}})=\mathbb{E}[h^2_{\mathbf{w}_{\textbf{O}}}({x})\mid y_{\textbf{O}}=-1] - b_{\textbf{O}}^2$, and $m_{\textbf{O}}$ is the margin that aims to push the expected prediction scores of negative and positive class far from each other \cite{yuan2020large}. Examining \eqref{eq:AUCLoss}, the minimization problem of $a_{\textbf{O}}$ and $b_{\textbf{O}}$ is achieved when $a_{\textbf{O}}=a(\mathbf{w}_{\textbf{O}})=\mathbb{E}[h_{\mathbf{w}_{\textbf{O}}}({x})\mid y_{\textbf{O}}=1]$, and $b_{\textbf{O}}=b(\mathbf{w}_{\textbf{O}})=\mathbb{E}[h_{\mathbf{w}_{\textbf{O}}}({x}^{\prime})\mid y_{\textbf{O}}=-1]$ respectively \cite{ying2016stochastic,yuan2020large}. Thus, we expect that minimizing \eqref{eq:AUCLoss} with respect to $\mathbf{w}_{\textbf{O}}$ can produce a well-tuned binary classifier for the imbalanced ``O''-tag prediction. Given $K$ labels, $K$ losses can be defined under OVA and independently minimized to produce an optimal binary classifier for each unique label.

At prediction time on the test data, we evaluate the individual classifiers by following the maximum confidence strategy \cite{GALAR20111761} to generate the entity tags expected by the corpus label set.

{
    \footnotesize
    \begin{equation}
        \hat{y_i} = \argmax_{1 \ldots K} \left[
         h_{\mathbf{w}^{*}_{1}}\left( x_i\right),
        \ldots,
         h_{\mathbf{w}^{*}_{K}}\left( x_i\right) 
        \right]
        ,
    \end{equation}
}where $\mathbf{w}^{*}_{k}$ represents the optimal parameters learnt by minimizing \eqref{eq:AUCLoss} for the $k$-label classifier and $h_{\mathbf{w}^{*}_{k}}\left( x_i\right)$ represents the probability that the $x_i$ token belongs to class $k$. We acknowledge that the maximum confidence strategy, although producing the appropriate label predictions, ignores the inherent weakness of OVA, {\it i.e.}, OVA can lead to confusion areas where {\bf (i)} two or more binary classifiers can be confident that the sample belongs to their classes, or {\bf (ii)} no classifiers are confident enough to claim the sample, especially under the imbalanced settings \cite{rifkin2004defense, liu2017action}. However, since AUC maximization can learn a well-tuned binary classifier under imbalanced settings, we argue that this weakness is less concerning, as shown in \autoref{fig:OVA_BCE_3d_comparison}. The results in the figure show that the NER model, trained with the AUC surrogate loss under the OVA setup, leads to smaller confusion areas (few samples in the lower left region) compared to the binary cross entropy loss, indicating that the binary classifiers, learnt with the AUC surrogate loss function, are more well-tuned than those learnt with the binary cross entropy loss. More statistical results and detailed discussions can be found in \autoref{sec:ExpRes}.

\begin{algorithm}[t]
    \caption{OVA-AUC}
    \label{alg:OVAAUC}
    \raggedright{
        \textbf{Input}: Training set $\mathcal{S}$, $\theta, \{ \omega_1,\ldots, \omega_K\}$, $\{a,b,\alpha\}^{K}$\\
        \textbf{Output}: $\theta^*, \{ \omega_1^*,\ldots, \omega_K^*\}$ \\
    }
    \begin{algorithmic}[1]
        \FOR{epoch in {\tt range}(num epochs)}
            \FOR{prefix in \{B,I, O\}}
                \STATE $\mathcal{L}:=0$
                \FOR{$i$ in {\tt range}(K)}
                    \IF {{\tt i.startswith}(prefix)}
                        \STATE $\mathcal{L}+= $ Equation~\eqref{eq:AUCLoss}
                    \ENDIF
                \ENDFOR
                \STATE $\mathcal{L}${\tt .optimize()}
            \ENDFOR
        \ENDFOR
    \end{algorithmic}
\end{algorithm}


\section{Experimental Settings}
\label{sec:ExpSett}
\textbf{Domains and Corpora:}
We used corpora from both general and biomedical domains to benchmark the NER performance. \autoref{tab:DatasetSetting_BIO} summarizes the label distribution statistics for these corpora. Both CoNLL 2003 \cite{tjong-kim-sang-de-meulder-2003-introduction} and OntoNotes5 \cite{weischedel2011ontonotes} are benchmark corpora from the general domains. Whereas NCBI \cite{10.5555/2598938.2599127}, and s800 \cite{10.1371/journal.pone.0065390} have been widely used in biomedical named entity recognition (bioNER) task \cite{10.1093/bioinformatics/btz682, nguyen-etal-2022-hardness}. Since these corpora vary in terms of label distribution and linguistic characteristics, they serve to substantiate the versatility of our NER methods regardless of the underlying corpora or domains.

\textbf{Model Architecture and Embedding:}
For embeddings and model architectures, we focused on state-of-the-art NER and bioNER architectures and embeddings. CoNLL 2003 and OntoNotes5 were trained with ``bert-base-cased'' \cite{devlin-etal-2019-bert}, while NCBI and s800 were trained with ``biobert-base-cased-v1.1'' \cite{10.1093/bioinformatics/btz682} to avoid out-of-vocabulary (OOV) issues in bioNER \cite{nguyen-etal-2022-hardness}.

\begin{algorithm}[t]
    \caption{OVA-AUC-MAML First Order Approximation}
    \label{alg:OVAAUCMAML}
    \raggedright{
        \textbf{Input}: Training set $\mathcal{S}$, $\theta, \{ \omega_1,\ldots, \omega_K\}$, $\{a,b,\alpha\}^{K}$\\
        \textbf{Output}: $\theta^*, \{ \omega_1^*,\ldots, \omega_K^*\}$ \\
    }
    \begin{algorithmic}[1]
        \FOR{epoch in {\tt range}(num epochs)}
            \STATE set $\mathcal{L}:=0$ and sample m classes
                \FOR{$i$ in {\tt range}(K)}
                    \IF {$i$ in m}
                        \STATE $\mathcal{L}+= $ Equation~\eqref{eq:AUCLoss}
                    \ENDIF
                \ENDFOR
            \STATE $\mathcal{L}${\tt .optimize()}
        \ENDFOR
    \end{algorithmic}
\end{algorithm}

\begin{figure*}[t]
    \centering
    \includegraphics[width=0.85\textwidth]{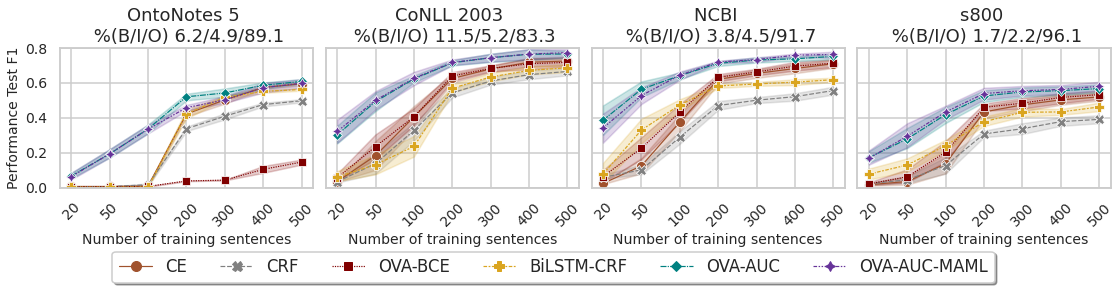}
    \caption{
        The average entity-F1 performance with 95\% error bands on the test set taken from 10 random training partitions of each training set $\mathcal{S}$ size for each loss function. The title of the plot indicates the corpus, and the label distribution in BIO format. 
    }
    \label{fig:all_corpora_sentences_test_f1}
\end{figure*}

\textbf{Baselines \& OVA AUC NER Methods:}
\begin{itemize}[topsep=0pt, partopsep=0pt, noitemsep, leftmargin=*]
    \item \textbf{CE}: The standard multi-class cross entropy loss, most commonly used in almost all existing NER works, was used as one of the major baselines to verify and establish the significance and impact of our OVA AUC NER methods.
    
    \item \textbf{CRFs}: As our second baseline, CRFs were traditionally used in many NER works, such as those of \cite{DBLP:journals/corr/LampleBSKD16,nguyen-etal-2022-hardness}. As CRFs produce a sequence labeller instead of a token classifier, we also used BiLSTM to push the performance of this baseline.
    
    \item \textbf{OVA-BCE}: Instead of using the AUC surrogate loss, this baseline simply uses the binary cross entropy loss (BCE) for each label. This baseline is to study the performance difference between BCE and the AUC surrogate loss under low-resource and imbalanced settings for ablation purposes.
    
    
    

    \item \textbf{OVA-AUC}: As OVA has higher training time compared to multi-class CE since each binary classifier should be independently trained, we thus consider grouping the binary classifiers for labels that share similar linguistic characteristics ({\it e.g.}, B-PER, B-ORG, B-MISC and B-LOC for CoNLL 2003) and train their classifiers together (see Algorithm~\ref{alg:OVAAUC}).

    \item \textbf{OVA-AUC-MAML}: Additionally, we also reduce the training time by applying first-order meta-learning (MAML) \cite{finn2017model,DBLP:journals/corr/abs-1803-02999} and sample a random batch of $m$ binary classifiers in each iteration for the model to optimize (see Algorithm~\ref{alg:OVAAUCMAML}).
\end{itemize}

It is noteworthy that the goal of our experiments is to compare the baseline objective functions and our OVA AUC objective functions; thus, to make the comparison fair, we used the same embeddings and language model for all the objective functions {\it e.g.}, Bert-CE, Bert-CRF, and Bert-BiLSTM-CRF v.s. Bert-OVA-AUC and Bert-OVA-AUC-MAML. 

\textbf{Evaluation Setups:}
\begin{itemize}[topsep=0pt, partopsep=0pt, noitemsep, leftmargin=*]
    \item The size of $\mathcal{S}$: In order to simulate the low-resource scenarios, we used training~set $\mathcal{S}^{}$ with size of $\{20,  50, 100, \allowbreak 200, 300, 400, 500\}$. We then trained our methods and all the baselines on 10 random training partitions of the same size and reported/analysed their average entity F1 performance.
    \item Imbalance entity generator: This was derived from \cite{nguyen2022auc} to sample $\mathcal{S}$ containing $\{1, 2, 5, 10\}$ percentage of entity tokens, {\it i.e.}, tokens with labels that are not ``O''. This is to investigate the robustness of our methods under different distributions.
\end{itemize}

\begin{figure*}[t]
    \centering
    \includegraphics[width=0.85\textwidth]{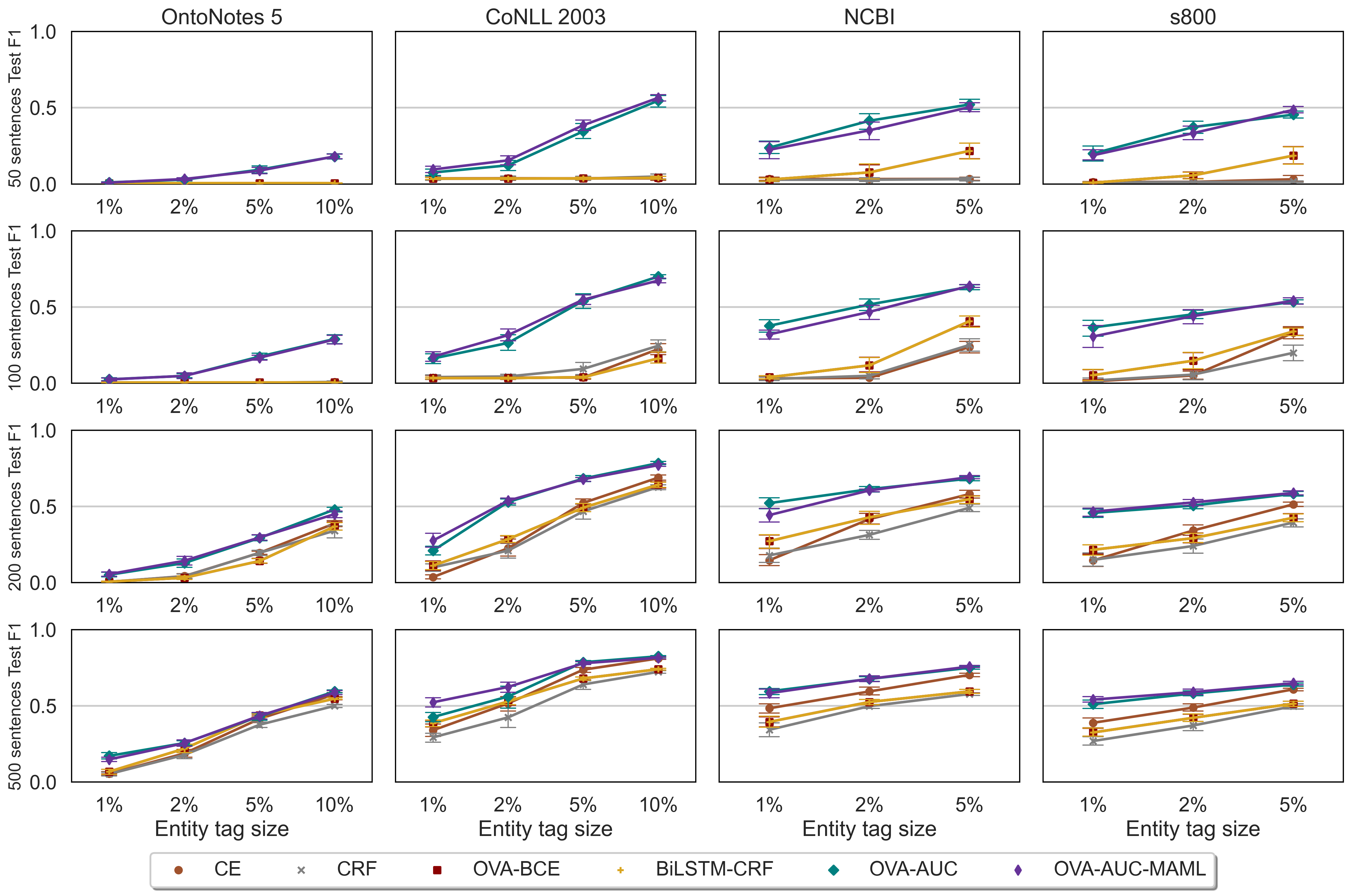}
    \caption{
        The average entity-F1 performance with error bands on test set taken from 10 random training partitions of each training set size for each loss function. The entity tag size indicates the percentage of entity-tokens to the total tokens in the training set. 
    } 
    \label{fig:four_50sent_imb_test_f1}
\end{figure*}

\section{Experimental Results \& Discussions}
\label{sec:ExpRes}
\subsection{Low-resource Studies}\label{sec:lowres}
Using \autoref{fig:all_corpora_sentences_test_f1}, we observe the followings:

\begin{itemize}[topsep=0pt, partopsep=0pt, noitemsep, leftmargin=*]
    \item {\bf CE vs. OVA-AUC}: Under extreme low-resource scenarios ({\it i.e.}, size $\{20, 50\}$), OVA-AUC outperforms CE by a significant margin, with the average performance difference reaching $30\%$. When the training pool increases, OVA-AUC still exhibits substantial gains compared to CE at the $95\%$ level of confidence for the vast majority of the time, indicating that OVA-AUC is a superior alternative to the standard multi-class CE loss function for low-resource NER.

    \item \textbf{CRF and BiLSTM-CRF vs. OVA-AUC}: It is apparent that OVA-AUC significantly outperforms CRF in all scenarios. As CRF is a sequence labeler, we further adopt BiLSTM to improve this baseline's performance. Nevertheless, \autoref{fig:all_corpora_sentences_test_f1} suggests
    that OVA-AUC should remain a superior solution to both CRF and BiLSTM-CRF under the low-resource NER.
    
    \item \textbf{OVA-BCE vs. OVA-AUC}: The performance of OVA-BCE is close to that of CE for CoNLL 2003, s800, and NCBI datasets. However, its performance is sub-optimal for the Ontonotes 5 dataset, resulting in a noticeable gap compared to our OVA-AUC training strategies. This difference can be attributed to the higher number of classes in the Ontonotes 5 dataset ($K=37$), which leads to an extremely imbalanced situation. One of the aforementioned weaknesses of the OVA approach is its inefficiency in learning from imbalanced data, often resulting in less accurate classifiers compared to multi-class counterparts. Therefore, these findings underscore the importance of combining both the OVA approach and AUC maximization to achieve optimal results for imbalanced NER.
    
    \item \textbf{OVA-AUC vs. OVA-AUC-MAML}: Although the performance of OVA-AUC can be higher on average than that of OVA-AUC-MAML under some settings, the difference between the two approaches is not significant, except for OntoNotes5 with a 200 training size. As OVA-AUC groups the labels that share similar linguistic characteristics and trains their classifiers together, the difference can be attributed to longer training time. As OVA-AUC-MAML performs on par with OVA-AUC, it aptly outperforms all the selected baselines in most low-resource NER across all the corpora.
\end{itemize}

\subsection{Imbalanced Data Distribution Studies}
We deployed an imbalance entity tag generator \cite{nguyen2022auc} to demonstrate the robustness of our methods on the diverse training set distributions for the NER task. This generator simulates scenarios in which the training set $\mathcal{S}$ data distribution changes from that of $\mathcal{S}^{\text{test}}$ in order to test the resilience of the baselines and our methods. Figure 3 illustrates the performance differences for those methods according to the size of the entity tag. From these results, we provide the following observations:
\begin{itemize}[topsep=0pt, partopsep=0pt, noitemsep, leftmargin=*]
    \item \textbf{CE vs. OVA-AUC}: Across all entity tags settings, we observed that OVA-AUC consistently outperforms CE. On the basis of the performance on CoNLL, NCBI, and s800, OVA-AUC is significantly superior to CE in the most extreme imbalanced scenarios ({\it i.e.}, entity label size of $1$ and $2\%$). As the amount of entity tags rises in OntoNotes5, OVA-AUC performance improves greatly. Conversely, CE performs inadequately when the entity tag and training size are small.
    
    \item \textbf{CRF and BiLSTM-CRF vs. OVA-AUC}: OVA-AUC outperforms CRF and BiLSTM-CRF by a significant margin in all scenarios. Similar to CE, neither CRF nor BiLSTM-CRF performs when the entity tag size is extremely small as the sequence labelers are not fed with enough learning signals.
    
    \item \textbf{OVA-BCE vs. OVA-AUC}: The results further reinforce the previous discussion that AUC trained binary classifiers under OVA are important to the low-resource and imbalanced NER.
    
    \item \textbf{OVA-AUC vs. OVA-AUC-MAML}: We observe no significant difference between OVA-AUC and OVA-AUC-MAML based on the performance in most settings across all datasets.
\end{itemize}

Overall, while the results from \autoref{sec:lowres} suggests that OVA AUC NER methods are important under the low-resource NER, this subsection brings new evidence to the versatility of our works under different NER imbalanced label distributions.

\section{Conclusion}
\label{sec:conclu}
In this study, we offer efficient approaches to the challenges of low-resource and imbalanced data that are ubiquitous in many NER tasks via restructuring the conventional NER multi-class learning problem into a one-vs-all learning scenario, and utilizing an AUC loss to instruct the binary classifiers. Our experiments on diverse datasets in low-resource and data imbalance situations prove the superiority of our OVA AUC NER strategies over traditional methods like CE and CRF, regardless of the NER models, embeddings, or domains being employed. 
These findings not only present advancements in handling low-resource and imbalanced NER but also pave the way for future exploration, with potential ramifications in wider NLP areas grappling with data imbalance and resource scarcity.

\bibliographystyle{IEEEtran}
\bibliography{IEEEabrv,mybibfile}
\end{document}